\documentclass[runningheads]{llncs}

\usepackage{eccv}

\usepackage{eccvabbrv}

\usepackage{graphicx}
\usepackage{booktabs}

\usepackage[accsupp]{axessibility}  

\definecolor{ego_blue}{HTML}{216B93}
\definecolor{ngb_green}{HTML}{A2DDB1}

\usepackage[hidelinks]{hyperref}
\usepackage{orcidlink}

\usepackage{graphicx}
\usepackage{amsmath,amssymb}
\usepackage{booktabs}
\usepackage{multirow}
\usepackage{hyperref}
 
\begin{document}
 
\title{One Diffusion Model, Two Roles: Guided Trajectory Planning and Safety-Critical Scenario Generation in Closed-Loop Simulation}
 
\titlerunning{Guided Diffusion for Planning and Scenario Generation}
 
\author{Arka Pal\inst{*, 1}\orcidlink{0009-0008-5800-3550} \and
  Rajesh Kumar\inst{*, 1}\orcidlink{0009-0000-4158-5736} \and Hannes Eriksson\inst{2} \and Rémi Lacombe\inst{2} \and Arvid Laveno Ling\inst{2,3} \and Ankit Gupta\inst{2} \orcidlink{0000-0002-5628-3623} \and Maciej Wozniak \inst{1,2} \orcidlink{0000-0002-3432-6151}  } 
\institute{
  KTH Royal Institute of Technology, Sweden\\
  \email{\{arkap, rajeshk, maciejw\}@kth.se}
  \and
  Zenseact AB, Sweden\\
  \email{\{hannes.eriksson, remi.lacombe, ankit.gupta\}@zenseact.com}
  \and
  Chalmers University of Technology, Sweden\\
  \email{laveno@chalmers.se} \\
  $^*$Equal contribution, Master's thesis performed at Zenseact.\\
}
\authorrunning{A. Pal et al.}
 
\maketitle
 
\begin{abstract}
Diffusion probabilistic models can capture the multi-modal, interaction-rich distribution of joint future trajectories in driving scenes. We show that a \emph{single} pretrained diffusion traffic model can serve two complementary roles in the autonomous driving development loop: as an ego motion planner, and as a controllable generator of safety-critical scenarios for stress-testing the planners. On the planning side, we introduce a Single-Stream Dual-Stream (SSDS) diffusion-transformer decoder that fuses scene context via joint attention rather than late cross-attention, improving closed-loop performance on nuPlan. We further propose Decoupled Annealing Posterior Sampling with Energy (DAPSE), a training-free guidance scheme that injects arbitrary energy functions (e.g., target speed) at the clean-sample level, avoiding the first-order approximation errors of diffusion posterior sampling while requiring no auxiliary networks. Beyond planning, we leverage the same diffusion model as a controllable scenario generator to create realistic long-tail driving interactions for closed-loop evaluation. Through inference-time guidance, selected agents are steered toward safety-critical behaviors, including aggressive cut-ins, lead-vehicle braking, and combined longitudinal-lateral interactions, while preserving realistic traffic behaviors. Evaluated in closed-loop nuPlan simulations with independent black-box planners, the generated scenarios expose failure modes that remain hidden under standard benchmarks. In particular, evaluated planners frequently rely on reactive braking responses rather than proactive evasive maneuvers when facing complex multi-agent interactions, revealing a limitation of current learned planners. Although the SSDS-based planner achieves stronger nominal performance, it experiences larger degradation under these challenging scenarios, demonstrating that benchmark superiority does not necessarily translate to robustness. These results demonstrate that a single learned traffic prior can simultaneously improve motion planning and provide a realistic framework for systematic planner robustness evaluation.

\end{abstract}
 
\section{Introduction}
\label{sec:intro}
 
Learning-based planners for autonomous vehicles must satisfy two requirements that are usually studied separately: they must produce safe, human-like trajectories in interactive traffic \cite{human_like_driving}, and they must be \emph{evaluated} against the rare, safety-critical interactions that the real world has to offer \cite{ding2023survey}. Imitation-based planners trained on expert demonstrations struggle with the first requirement because standard regression objectives tend to average over the multi-modal set of plausible behaviors \cite{imitation_average, diffusion_planner}; standard benchmarks struggle with the second because logged data is dominated by nominal driving, so strong benchmark scores do not imply robustness in long-tail interactions \cite{nureasoning}.
 
Diffusion Probabilistic Models (DPMs)\cite{ddpm} can address both problems with the same mechanism. By learning the joint distribution of future trajectories of the ego vehicle and its neighbors conditioned on scene context, a DPM (i) captures multi-modality natively, and (ii) exposes an inference-time control surface: the reverse process can be steered by gradients of arbitrary energy functions without retraining. Steering the \emph{ego} towards low-energy (safe, comfortable, rule-compliant) trajectories yields a planner; steering a \emph{neighbor} towards high-risk interactions with the ego yields an adversarial scenario generator. 

\begin{figure}[!ht]
  \begin{center}
    \includegraphics[width=\textwidth]{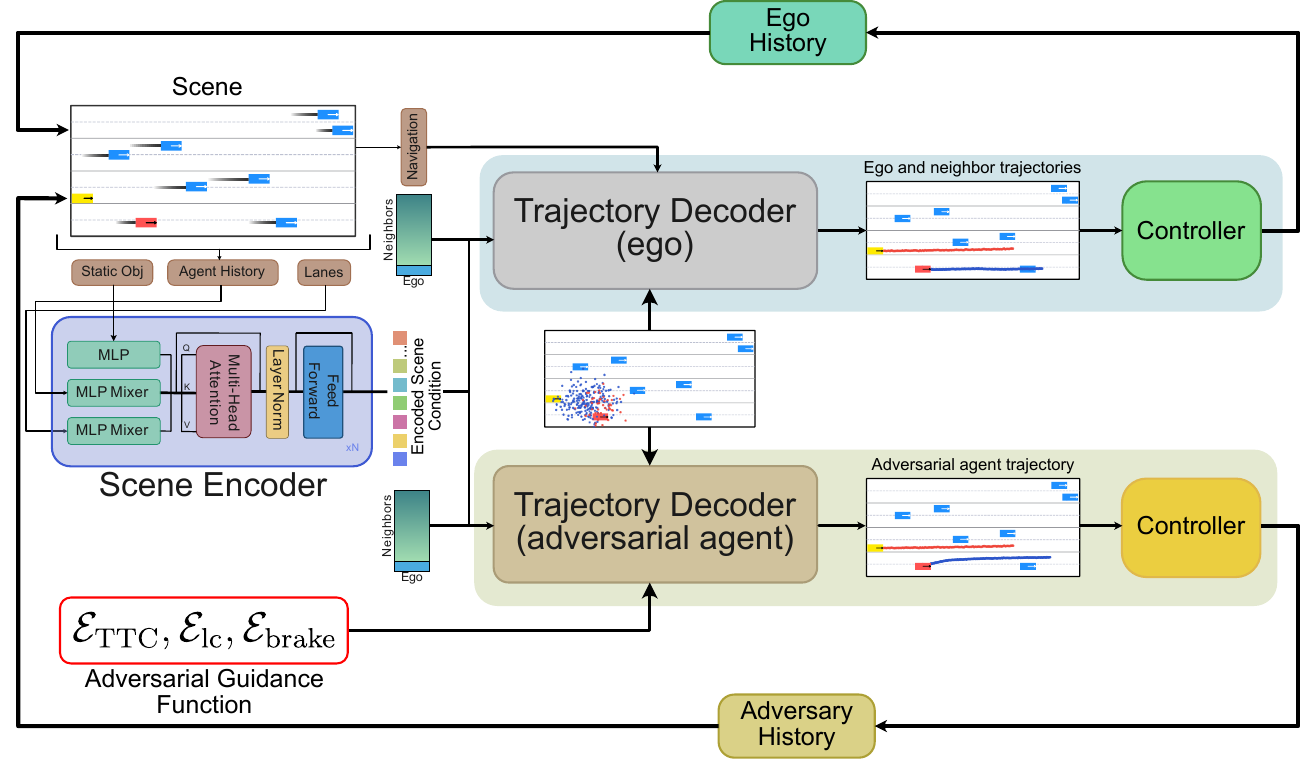}
  \end{center}
  \caption{The dual planner and scenario generator framework. A shared scene encoder produces a scene condition used by two trajectory decoders: the ego decoder and the adversarial decoder are sampled independently, with the adversarial decoder being steered at each denoising step by the gradient of a scenario-specific energy (e.g. $\mathcal{E}_{\mathrm{TTC}}, \mathcal{E}_{\mathrm{lc}}, \mathcal{E}_{\mathrm{brake}}$). Both trajectories are tracked by an LQR controller, and the resulting histories are fed back for the next timestep.}
  \label{fig:dual_framework}
\end{figure}

 
This paper develops both directions, as presented in Figure \ref{fig:dual_framework}, on top of a Diffusion Transformer (DiT)-based \cite{dit} traffic model trained on nuPlan \cite{nuplan}, an autonomous driving planning benchmark with 1500 hours of human driving data, and makes the following contributions:
 
\begin{enumerate}
\item \textbf{SSDS-DiT decoder for joint prediction and planning.} We replace the standard DiT decoder with a Single-Stream Dual-Stream architecture adapted from latest video generation models. Dual-stream blocks process trajectory and context tokens in separate streams that interact through joint attention; a subsequent single-stream stage fuses the unified representation. 

\item \textbf{DAPSE: Training-free exact guidance for zero-shot trajectory steering.} To bypass the costly retraining of existing safety-steering methods, we propose \emph{Decoupled Annealing Posterior Sampling with Energy} (DAPSE), enabling zero-shot enforcement of driving constraints on our pre-trained diffusion-based planner.
\item \textbf{Controllable safety-critical scenario generation in closed loop.} We repurpose the same pretrained diffusion model as a behavior generator for a designated adversary within a fully closed-loop nuPlan simulation. We generate controllable and realistic safety-critical highway interactions, including aggressive cut-ins and sudden braking events, where the adversary challenges the ego planner through targeted driving behaviors.

\end{enumerate}

\section{Related Work}
\label{sec:related}
 
\textbf{Data-Driven Motion Planning} Traditional local planners rely heavily on hand-crafted rules, sampling, or Model Predictive Control (MPC), which struggle to scale to dense, unpredictable urban environments due to local optima and rigid parameter tuning \cite{mpc3}. To scale motion planning to these complex settings, Imitation Learning (IL) has emerged as a highly effective paradigm \cite{IL_org}, implicitly learning the rules of the road from large-scale expert datasets: either with end-to-end pipelines \cite{uniad, prix} or via object-level representations \cite{pluto}. However, standard regression-based IL is plagued by mode averaging, where a single deterministic output fails to capture the diverse, multi-modal behaviors inherent in human driving \cite{diffusion_planner}. This limitation often necessitates large-scale, computationally expensive reinforcement learning rollouts \cite{gigaflow} or rule-based fallback safety filters \cite{safetynet} to reject unsafe trajectories.
 
\noindent \textbf{Diffusion models for trajectories.} DPMs have emerged as a powerful paradigm for capturing highly multi-modal, complex trajectory distributions without the risks of mode averaging. Early frameworks introduced trajectory diffusion to robotic manipulation by directly synthesizing state-action sequences \cite{MSE}. This was subsequently extended to autonomous driving for multi-agent motion forecasting tasks, demonstrating superior capability in predicting diverse future intents compared to traditional regression or generative adversarial models \cite{motiondiffuser}.

To unify upstream prediction with downstream planning, recent architectures scale this approach to joint trajectory generation. For instance, Gen-Drive \cite{gen_drive} leverages a behavior diffusion model as a multi-agent scene generator to simulate interactive future scenarios. Furthermore, to capture true multi-agent interactions, Large Trajectory Models (LTMs) \cite{LTM} unify prediction and planning within massive, data-intensive transformer structures.

Most relevant to our work, Diffusion Planner \cite{diffusion_planner} utilizes a Diffusion Transformer (DiT) \cite{dit} decoder to model scene interactions natively within the generative process. By jointly processing past agent histories to output both the ego-vehicle's plan and neighboring agent trajectories, it serves as an expressive, interactive foundation for controllable driving behavior. 

\noindent \textbf{Safety-critical scenario generation.}
Automatically generating realistic driving scenarios has become an important direction for scalable evaluation of motion planners. Recent generative approaches, including TrafficGen~\cite{feng2022trafficgen}, LCTGen~\cite{tan2023language}, and ProSim~\cite{tan2024promptable}, demonstrate the ability to synthesize diverse and controllable traffic behaviors. However, evaluating planner robustness requires scenarios beyond nominal traffic distributions, motivating adversarial scenario generation methods such as AdvSim~\cite{Wang2021AdvSim}, which optimizes adversary behaviors to induce planner failures but may produce unrealistic trajectories. To improve realism, STRIVE~\cite{rempe2022strive} introduces a learned traffic prior that captures joint multi-agent behavior. By constraining the optimization to the learned traffic manifold, the generated scenarios preserve more realistic interactions while remaining safety-critical. However, it provides limited control over the generated scenario and performs scenario generation in an open-loop setting. 

More recently, SafeSim~\cite{chang2024safesim} adopts a diffusion-based traffic model with inference-time guidance to enable controllable and reactive safety-critical scenario generation in closed-loop simulation. However, it primarily focuses on intersection-based scenarios and relies on trajectory proposals to guide adversarial behaviors. Building upon diffusion-based scenario generation, our work targets high-speed highway interactions, where limited reaction time makes aggressive cut-ins and sudden braking particularly challenging for motion planners. Instead of prescribing adversarial trajectories, the diffusion model itself generates multi-modal traffic behaviors, while interpretable energy-based guidance steers the generated trajectories toward realistic safety-critical interactions.



In our work, we build directly upon the joint prediction-and-planning formulation, with a more expressive decoder that effectively fuses the scene condition during trajectory generation. To evaluate the above-mentioned planner under challenging long-tail scenarios, we leverage the same diffusion framework to generate controllable adversarial scenarios. Through energy-based guidance, generated agent behaviors are steered toward realistic safety-critical interactions.


\section{Preliminaries: Diffusion over Joint Trajectories}
\label{sec:prelim}
 
Let the scene condition $\mathbf{E_f}$ collect current and historical agent states, lane polylines with traffic-light status, static objects, and the ego navigation route. The generated trajectory representation is represented as a state tensor $\boldsymbol{x}^{(t)} \in \mathbb{R}^{(M+1) \times \tau \times 4}$ at diffusion time step $t \in [0, 1]$. An element of this tensor, ${x}_{i, k}^{(t)}$, represents the state of agent $i \in \{0, 1, \dots, M\}$ at trajectory time step $k \in \{1, \dots, \tau\}$, where $i = 0$ denotes the ego vehicle and $i > 0$ denotes its $M$ nearest neighbors. 

The clean target state ($t=0$) for agent $i$ at trajectory step $k$ is given by:
\begin{equation}
{x}_{i, k}^{(0)} = \big[\, p_{x, i}^k, \, p_{y, i}^k, \, \cos\theta_i^k, \, \sin\theta_i^k \,\big],
\end{equation}
where $p_{x, i}^k$ and $p_{y, i}^k$ denote the $x$ and $y$ coordinates, and $\theta_i^k$ is the heading angle. 

A variance-preserving forward process with a linear noise schedule corrupts the clean trajectory tensor $\boldsymbol{x}^{(0)}$ to Gaussian noise $\boldsymbol{x}^{(1)}$. The model $\mu_\theta(\boldsymbol{x}^{(t)}, t, \mathbf{E_f})$ is trained to predict the clean sample from noisy data \cite{ramesh_bhai}.
 
\section{Planner: SSDS Diffusion Planning with DAPSE Guidance}
\label{sec:planning}
 
\subsection{Scene encoder}
The driving environment consists of various multi-modal data, namely lane information, traffic light status, navigation routes, agent history, static objects, etc. Fusing this spatio-temporal information is important for the downstream trajectory planning task. Inspired by the latest success of diffusion-based trajectory planning \cite{diffusion_planner}, we follow the below-mentioned scene encoder (\textit{Scene Encoder} block in Figure \ref{fig:dual_framework}): agent histories and lane features are processed by per-modality MLP-Mixer \cite{mlpmixer} blocks to mix independently across feature and sequence dimensions; static objects are processed by an MLP; the concatenated tokens pass through self-attention and feed-forward layers to yield the fused context $\mathbf{E_f}$. The navigation route is encoded separately and added to the diffusion-time embedding to form the conditioning vector for adaptive layer norm (adaLN) \cite{dit}.


\subsection{Single-Stream Dual-Stream (SSDS) decoder}
\label{sec:ssds}
\begin{figure}[!ht]
  \begin{center}
    \includegraphics[width=\textwidth]{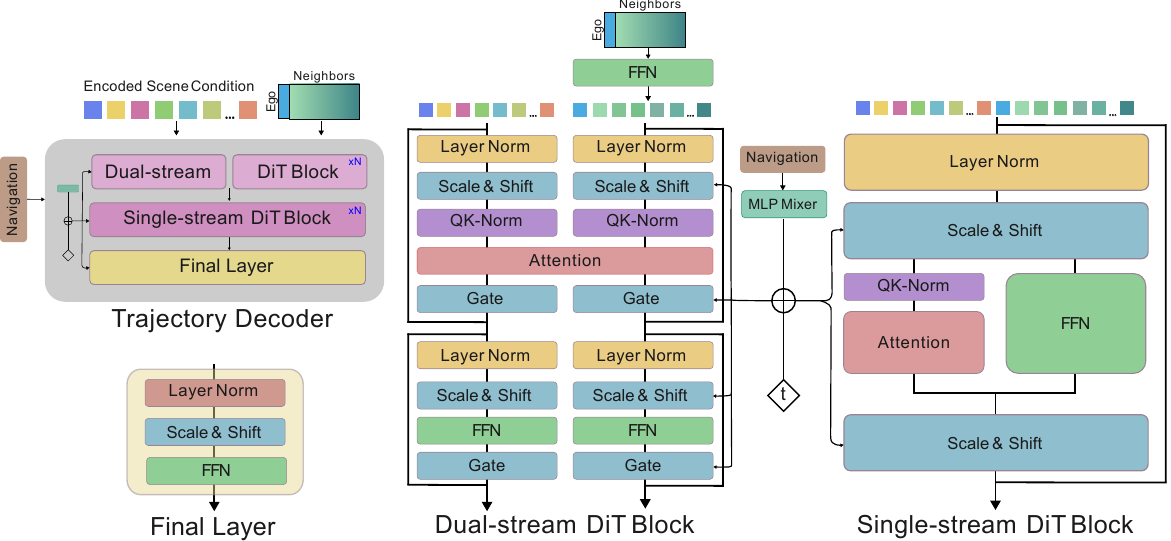}
  \end{center}
  \caption{The proposed Single-stream and Dual-stream based DiT architecture for planning of \textcolor{blue}{ego} and prediction of \textcolor{teal}{neighbors}.}
  \label{fig:decoder}
\end{figure}

To effectively fuse scene context and trajectories in the decoder, we take motivation from multi-modal video-generation architectures \cite{videogen_ssds, epona}. Our decoder treats trajectory tokens $\boldsymbol{x}^{(0)}$ and context tokens $\mathbf{E_f}$ as two streams, as presented in Figure \ref{fig:decoder}. In each \emph{dual-stream} block, both streams receive independent adaLN modulation and QKV projections \cite{transformer}; queries, keys, and values are concatenated for a single \emph{joint attention} operation. A subsequent \emph{single-stream} stage concatenates the streams into a unified token set processed by parallel attention and MLP branches. All gates, scales, and shifts are conditioned on the navigation and diffusion time-step embedding, so route information modulates every layer.  With this setup, dual-stream blocks enable early, symmetric cross-stream interaction while preserving modality-specific processing, and the single-stream stage performs deep fusion---in contrast to the asymmetric, late fusion of cross-attention in standard DiT architecture \cite{dit}.

\subsection{DAPSE: Decoupled Annealing Posterior Sampling with Energy}
\label{sec:dapse}
Once a diffusion model is trained to capture the trajectory distribution $q_0(\boldsymbol{x}^{(0)})$, it can be further utilized to draw samples from a target distribution $p_0(\boldsymbol{x}^{(0)})$ that minimizes a given energy function $\mathcal{E}_0(\boldsymbol{x}^{(0)})$ via the following formulation \cite{exponential_tilting}:

\begin{equation}
p_0(\boldsymbol{x}^{(0)}) \;\propto\; q_0(\boldsymbol{x}^{(0)})\, e^{-\beta\, \mathcal{E}_0(\boldsymbol{x}^{(0)})}.
\label{eq:tilted}
\end{equation}
$\beta \geq 0$ is the inverse temperature that controls the energy strength. Sampling from $p_0(\boldsymbol{x}^{(0)})$ requires the corresponding score function $\nabla_{\boldsymbol{x}^{(t)}} \log p_t(\boldsymbol{x}^{(t)})$. However, we only have access to $\nabla_{\boldsymbol{x}^{(t)}} \log q_t(\boldsymbol{x}^{(t)})$ and the relation between the score functions is $\nabla_{\boldsymbol{x}^{(t)}} \log p_t(\boldsymbol{x}^{(t)}) =
\nabla_{\boldsymbol{x}^{(t)}} \log q_t(\boldsymbol{x}^{(t)})
\;-\;
\nabla_{\boldsymbol{x}^{(t)}} \mathcal{E}_t(\boldsymbol{x}^{(t)})$ \cite{exact_energy_guide}. Because the exact intermediate guidance $\nabla_{\boldsymbol{x}^{(t)}}\mathcal{E}_t(\boldsymbol{x}^{(t)})$ is intractable, existing methods typically follow one of two approaches. They either train auxiliary networks~\cite{exact_energy_guide, MSE}, which violate the strict real-time latency constraints ($10\text{--}20\,$Hz) of vehicle planners, or rely on test-time approximations such as Diffusion Posterior Sampling (DPS)~\cite{DPS}, which are typically reliable only near $t \approx 0$. To address these limitations, we propose \emph{Decoupled Annealing Posterior Sampling with Energy} (\emph{DAPSE}). Our method builds upon DAPS \cite{DAPS} by generalizing its consecutive noise-level decoupling framework from standard measurement likelihoods to arbitrary, non-analytical energy functions.

Following the non-Markovian forward process framework of DDIM \cite{ddim} and decoupled sampling scheme of DAPS \cite{DAPS} ($\boldsymbol{x}^{(t_1)} \perp \!\!\! \perp \boldsymbol{x}^{(t_2)} \mid \boldsymbol{x}^{(0)}$) , we can express the transition probability by marginalizing over the clean data manifold:
\begin{align}
    p_{t_2}(\boldsymbol{x}^{(t_2)}) 
    &= \int p_{t_1}(\boldsymbol{x}^{(t_1)})p(\boldsymbol{x}^{(0)}| \boldsymbol{x}^{(t_1)}) p(\boldsymbol{x}^{(t_2)}| \boldsymbol{x}^{(0)}, \boldsymbol{x}^{(t_1)})d\boldsymbol{x}^{(0)}d\boldsymbol{x}^{(t_1)} \nonumber\\
    &= \int p_{t_1}(\boldsymbol{x}^{(t_1)})p(\boldsymbol{x}^{(0)}| \boldsymbol{x}^{(t_1)}) p_{t_20}(\boldsymbol{x}^{(t_2)}| \boldsymbol{x}^{(0)})d\boldsymbol{x}^{(0)}d\boldsymbol{x}^{(t_1)} \nonumber\\
    &= \mathbb{E}_{\boldsymbol{x}^{(0)} \sim p(\boldsymbol{x}^{(0)} \mid \boldsymbol{x}^{(t_1)})}
       \mathcal{N}(\boldsymbol{x}^{(t_2)};\, \boldsymbol{x}^{(0)},\, \sigma_{t_2}^2 \boldsymbol{I})
\end{align}
The last line can be written as a known sample $\boldsymbol{x}^{(t_1)}$ from energy-conditioned $p(\boldsymbol{x}^{(t_1)})$ will be available from the previous iteration, and $p(\boldsymbol{x}^{(t_1)} \mid \boldsymbol{x}^{(0)})$ is defined by the standard forward Gaussian corruption process. To draw the energy-conditioned clean sample $\boldsymbol{x}^{(0)} \sim p(\boldsymbol{x}^{(0)} \mid \boldsymbol{x}^{(t)})$, we factorize the posterior using Bayes' rule and reuse the formulation in Eq.~\eqref{eq:tilted}:
\begin{align}
p(\boldsymbol{x}^{(0)} \mid \boldsymbol{x}^{(t)}) 
&\propto p(\boldsymbol{x}^{(t)} \mid \boldsymbol{x}^{(0)}) \, p_0(\boldsymbol{x}^{(0)}) \nonumber \\
&= q(\boldsymbol{x}^{(t)} \mid \boldsymbol{x}^{(0)}) \, q_0(\boldsymbol{x}^{(0)}) \, e^{-\beta \mathcal{E}_0(\boldsymbol{x}^{(0)})} \nonumber \\
&\propto q(\boldsymbol{x}^{(0)} \mid \boldsymbol{x}^{(t)}) \, e^{-\beta \mathcal{E}_0(\boldsymbol{x}^{(0)})},
\label{eq:factorization}
\end{align}
where we exploit the fact that the forward corruption likelihood $p(\boldsymbol{x}^{(t)} \mid \boldsymbol{x}^{(0)})$ is identical to the unguided $q(\boldsymbol{x}^{(t)} \mid \boldsymbol{x}^{(0)})$ \cite{exact_energy_guide}. 

To sample from this unnormalized distribution, we utilize Langevin dynamics \cite{langevin} over inner MCMC \cite{MCMC} iterations $j = 0, \dots, J-1$ with step size $\eta$:
\begin{equation}
\boldsymbol{x}^{(0,\,j+1)} = \boldsymbol{x}^{(0,\,j)} + \eta \nabla_{\boldsymbol{x}^{(0,\,j)}} \log p(\boldsymbol{x}^{(0,\,j)} \mid \boldsymbol{x}^{(t)}) + \sqrt{2\eta}\,\boldsymbol{\epsilon}^{(j)}.
\end{equation}

By approximating the intractable unguided posterior as a Gaussian centered around the current reverse ODE trajectory estimate $\hat{\boldsymbol{x}}^{(0)}(\boldsymbol{x}^{(t)})$ with heuristic variance $r_t^2 \boldsymbol{I}$ \cite{DAPS}, i.e., $q(\boldsymbol{x}^{(0)} \mid \boldsymbol{x}^{(t)}) \approx \mathcal{N}(\boldsymbol{x}^{(0)}; \hat{\boldsymbol{x}}^{(0)}(\boldsymbol{x}^{(t)}), r_t^2 \boldsymbol{I})$, the update step simplifies to our final DAPSE formulation:
\begin{equation}
    \boldsymbol{x}^{(0,\,j+1)}
    = \boldsymbol{x}^{(0,\,j)}
    \;-\;
    \underbrace{\eta \nabla
    \frac{\left\lVert \boldsymbol{x}^{(0,\,j)} - \hat{\boldsymbol{x}}^{(0)}(\boldsymbol{x}^{(t)}) \right\rVert^2}
         {2 r_t^{\,2}}}_{\text{Reconstruction term}}
    \;-\;
    \underbrace{\eta \beta \nabla \mathcal{E}_0\!\left( \boldsymbol{x}^{(0,\,j)} \right)}_{\text{Energy guidance}}
    \;+\;
    \underbrace{\sqrt{2\eta}\,\boldsymbol{\epsilon}^{(j)}}_{\text{Langevin noise}} 
\label{eq:dapse}
\end{equation}

This decoupling allows the sampler to correct global errors committed at early steps. Crucially, setting $\mathcal{E}_0 = -\log q_0(y \mid \boldsymbol{x}^{(0)})$ recovers the original DAPS update \cite{DAPS}, rendering Eq.~\eqref{eq:dapse} a strict generalization.

\section{Scenario Generator: Adversarial Scenario Generation}
While robust planning is essential for autonomous driving, its performance must also be evaluated under rare and safety-critical interactions that are underrepresented in nominal driving datasets. To assess planner robustness beyond in-distribution scenarios, we repurpose the pretrained DiT decoder for adversarial scenario generation through guided sampling. While DAPSE is our primary contribution for high-precision ego planning, we adopt DPS \cite{DPS} for adversarial scenario generation. This choice enables testing of diverse, composable adversarial energy functions without the added hyperparameter overhead of DAPSE’s annealing schedules. 
\label{sec:scenario}
 
\subsection{Closed-loop pipeline}

We integrate our guided diffusion framework into the nuPlan closed-loop simulation environment for interactive adversarial driving evaluation (Figure \ref{fig:dual_framework}). The simulation includes four agent categories: (1) the ego vehicle controlled by an independent black-box planner, (2) a manually designated adversary, whose behavior is generated by our guided diffusion model, (3) background vehicles following the Intelligent Driver Model (IDM) \cite{idm}, and (4) non-vehicle agents following log playback. At each timestep, the framework receives HD map information and agent histories to generate joint predictions, enabling the adversary to adapt to ego behavior. Guidance is selectively applied to the adversary’s positional trajectory during denoising, with headings recomputed from the resulting waypoints for geometric consistency. The guided trajectories are tracked using an LQR controller with a kinematic bicycle model \cite{polack2017kinematic} to generate executable control commands. As the ego vehicle reacts dynamically, safety-critical scenarios emerge through closed-loop interaction rather than predefined collision scripts.

\subsection{Guidance design principles}


Following the DPS formulation, guidance is applied during the low-noise stage of the reverse diffusion process, where trajectory refinement is most effective. Prior work using DPS guidance~\cite{diffusion_planner} performs denoising with 10 diffusion steps, which limits the number of guidance updates. To provide more opportunities for trajectory refinement, we increase the denoising steps to 20 and adopt a logSNR-based timestep schedule \cite{lu2025dpm} that allocates more steps to the low-noise region ($t \in [8{\times}10^{-4},0.1]$). To efficiently perform sampling, we use DPM-Solver ++~\cite{lu2025dpm} as the reverse diffusion solver, enabling stable guided sampling with fewer solver evaluations while maintaining sample quality.

The effectiveness of gradient-based guidance depends on the design of the guidance objective and the quality of the resulting gradients. Following the principles discussed in~\cite{diffusion_planner}, we design energy functions that produce smooth and stable guidance signals for trajectory refinement. We apply guidance directly to the positional trajectory rather than higher-order states such as velocity or acceleration, as position-level guidance provides a more stable way to influence the resulting motion. This allows higher-order behaviors, such as braking and evasive maneuvers, to emerge naturally from the generated trajectory dynamics.


\subsection{Guidance Function for long-tail scenarios}
\textbf{Time-to-Collision (TTC) with temporal lead.} We adopt the TTC objective used in prior safety-critical scenario generation methods such as STRIVE \cite{rempe2022strive} and SafeSim \cite{chang2024safesim}. Unlike purely distance-based objectives, TTC incorporates both the relative positions and velocities of interacting agents, providing a measure of future collision risk rather than instantaneous proximity alone. Following \cite{pmlr-v205-nishimura23a}, the TTC energy is defined as

\begin{equation}
\mathcal{E}_{\mathrm{TTC}} = \sum_{k=1}^{\tau}
- \exp\left(
- \frac{\bar{t}_{{\mathrm{col}}}(k)^2}{2 \lambda_t}
- \frac{\bar{d}_{{\mathrm{col}}}(k)^2}{2 \lambda_d}
\right),
\end{equation}

\noindent where $k$ denotes the predicted trajectory timestep over the planning horizon, and $\bar{t}_{\mathrm{col}}(k)$ and $\bar{d}_{\mathrm{col}}(k)$ represent the time and distance at closest approach under a constant-velocity assumption. The parameters $\lambda_t$ and $\lambda_d$ control the temporal and spatial sensitivity of the guidance. Directly minimizing this objective can encourage the adversary to target the ego vehicle's current position, often resulting in abrupt or unavoidable collisions that leave little opportunity for the ego planner to react. To encourage more plausible interactions, we introduce a temporal lead parameter $\ell$, whereby the TTC objective is evaluated between the adversary at trajectory timestep $k$ and the ego vehicle at the future trajectory timestep $k+\ell$, i.e., $(p_{\mathrm{adv}}(k), p_{\mathrm{ego}}(k+\ell))$ instead of $(p_{\mathrm{adv}}(k), p_{\mathrm{ego}}(k))$. This simple modification preserves the original TTC objective while shifting the interaction target from the ego vehicle's current position to its anticipated future position. As a result, the adversary is encouraged to intercept the ego's future trajectory, producing more temporally consistent and causally meaningful safety-critical interactions.

\noindent \textbf{Lane change (cut-ins).} To generate structured cut-in behaviors, we introduce a lane-change guidance objective. The guidance is activated through a one-shot trigger based on longitudinal distance. After activation, the adversary is encouraged to align with the ego lane centerline by minimizing the lateral deviation $\delta_{\mathrm{lat}}(k)=p^{\mathrm{adv}}_y(k)-c_{\mathrm{lat}}(k)$, where $p^{\mathrm{adv}}_y(k)$ is the adversary's lateral position and $c_{\mathrm{lat}}(k)$ is the corresponding lane-centerline lateral position in the ego-anchor frame. The lane-change energy is defined as

\begin{equation}
\mathcal{E}_{\mathrm{lc}}
=
\frac{1}{\tau}
\sum_{k=1}^{\tau}
\delta_{\mathrm{lat}}(k)^2 .
\end{equation}

\noindent Minimizing $\mathcal{E}_{\mathrm{lc}}$ drives the adversary toward the lane centerline, producing a lateral maneuver that follows the underlying road structure.
 
\noindent \textbf{Lead-vehicle braking.} The braking guidance objective is designed to induce smooth longitudinal deceleration of the adversary. The guidance is activated using a one-shot trigger based on the target speed and remains active after activation. Let $\mathbf{p}(k)$ denote the predicted position of the adversary and $\hat{\mathbf{h}}=(\cos\theta,\sin\theta)$ its current heading direction. The longitudinal displacement is computed as $s(k)=(\mathbf{p}(k)-\mathbf{p}(0))\cdot\hat{\mathbf{h}}$, with the desired progress defined as $s^*(k)=kv^*\Delta t$, where $v^*$ is the target velocity. The braking energy is defined as

\begin{equation}
\mathcal{E}_{\mathrm{brake}}
=
\frac{1}{\tau}
\sum_{k=1}^{\tau}
\left(\max(s(k)-s^*(k),0)\right)^2 .
\end{equation}

This objective penalizes longitudinal progress beyond the desired braking profile, encouraging the agent to gradually reduce its forward motion while maintaining lane consistency. The target velocity $v^*$ is progressively updated using a predefined deceleration rate until the desired minimum speed is reached, resulting in smooth and physically plausible braking behavior.

\noindent \textbf{Drivable-area compliance.} To maintain road compliance during guided trajectory generation, we incorporate a drivable-area constraint following the Euclidean Signed Distance Field (ESDF)-based formulation in PLUTO \cite{pluto}. The road geometry is represented by $\mathcal{D}(\mathbf{p})$, where positive values indicate drivable regions. Similar to \cite{pluto}, vehicle coverage circles are used to evaluate the distance field along the agent body. The drivable-area compliance energy is defined as

\begin{equation}
\mathcal{E}_{\mathrm{DA}}=
\frac{1}{\tau N_c}
\sum_{k=1}^{\tau}
\sum_{i=1}^{N_c}
\max(0,R_c-\mathcal{D}(\mathbf{c}_i(k))),
\end{equation}

\noindent where $\mathbf{c}_i(k)$ denotes the center of the $i$-th coverage circle, $R_c$ is its radius, and $N_c$ is the number of coverage circles used to approximate the vehicle footprint. Minimizing $\mathcal{E}_{\mathrm{DA}}$ penalizes off-road trajectories and encourages generated behaviors to remain consistent with the road geometry.
 
\noindent \textbf{Combined cut-in and braking}. Individual guidance objectives can be composed sequentially to generate more challenging interactions. For example, a lead vehicle that is initially too far ahead to produce a conflict can first be guided to brake, reducing the gap to the ego vehicle, before executing a cut-in maneuver. Depending on the initial traffic configuration, the order can also be reversed (cut-in followed by braking). The transition between objectives is controlled by a distance threshold $d_{\mathrm{comb}}$, enabling a wider range of coordinated longitudinal and lateral adversarial behaviors.
 
\section{Experiments}
\label{sec:experiments}
 
\subsection{Setup}
All models are trained on nuPlan \cite{nuplan} using different amounts of training data: (i) 250 thousand (250K); (ii) 650 thousand (650K); and (iii) 1 million (1M) extracted scenarios, and evaluated in closed loop on the Val14 \cite{pdm}, Test14, and Test14-hard splits \cite{rethinkingIL} in reactive (R) and non-reactive (NR) modes using the standard nuPlan score. We compare the reproduced baseline Diffusion Planner \cite{diffusion_planner} (DP) and our SSDS variant (\textbf{SSDS-DP}). For adversarial scenario generation, scenes and adversaries are mined from the Val14 split (1,118 scenarios across 14 types) to satisfy the geometric preconditions of each scenario type. Selection was manual but based on fixed agent-configuration criteria defined prior to and independent of either planner's evaluation (e.g., a nearby agent, ahead or behind, in an adjacent lane for lane-change, or a lead agent in the ego lane for braking). This process predominantly yielded scenes from the "following lane with lead" and "high-magnitude speed" categories, from which 25 scenes were used for the adversarial evaluation. As this process was manual, we cannot fully rule out an unintentional selection bias. Adversarial evaluation is performed only for the 1M-trained DP and SSDS-DP models using the proposed closed-loop pipeline.


\subsection{Planning: SSDS DP vs.\ DP}

\begin{figure}[!ht]
  \begin{center}
    \includegraphics[width=0.9\textwidth]{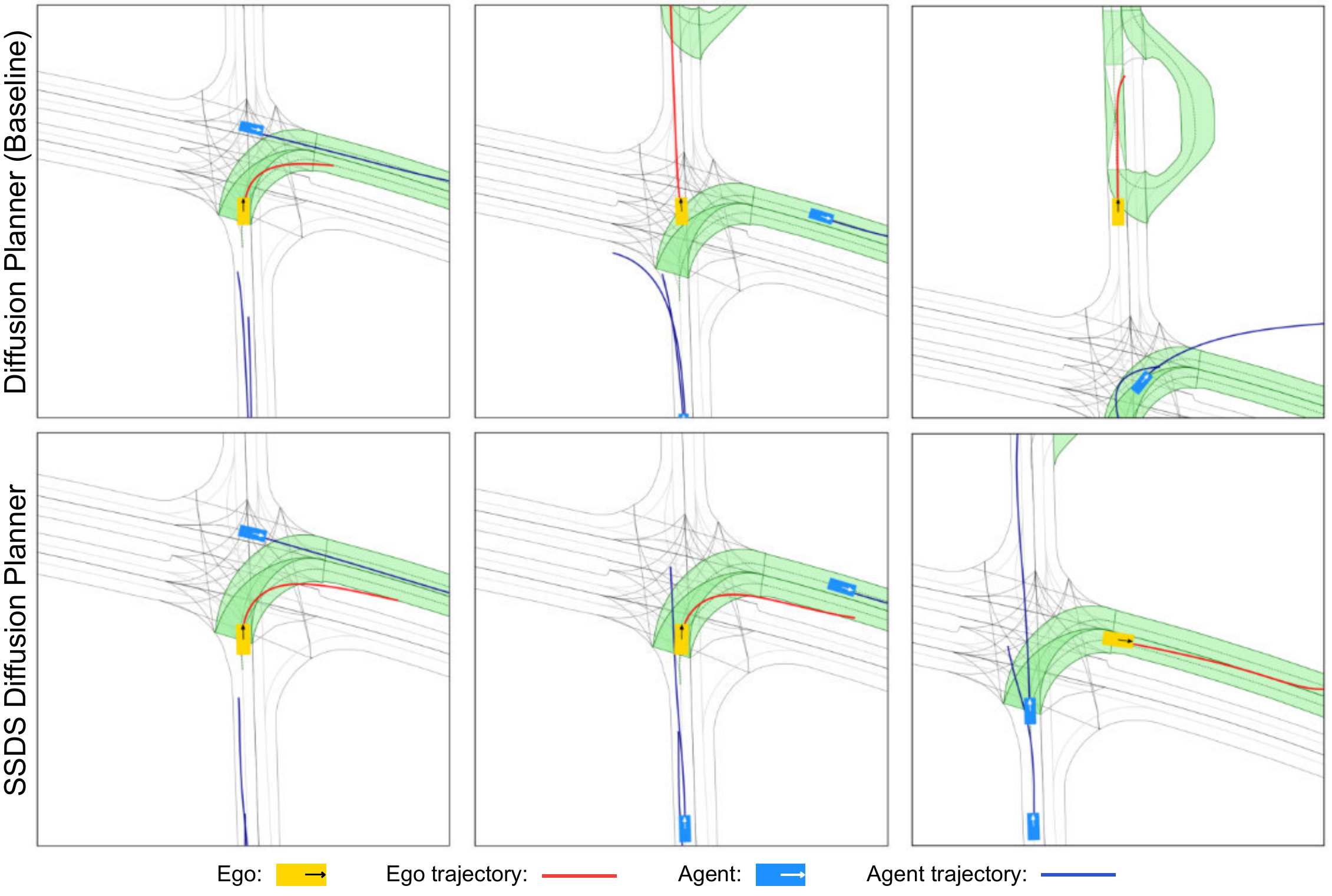}
  \end{center}
  \caption{Comparison between DP (baseline) and SSDS-DP (650K) in a Val14 Reactive simulation.}
  \label{fig:comp_navi}
\end{figure}

Comparative performances of DP~\cite{diffusion_planner} and \textbf{SSDS-DP} (ours) are presented in Table~\ref{tab:planning}, which reveals a few patterns. First, SSDS-DP improves most where interaction modeling matters: reactive, hard scenarios (Test14-hard Reactive: $+6.8$ at 250K, $+14.4$ at 650K). Second, the gap narrows in non-reactive settings, supporting the interpretation that joint attention chiefly benefits inter-agent dependency modeling rather than marginal trajectory accuracy. Third, SSDS-DP is \textit{more data efficient}, i.e., it consistently outperforms the baseline when models are trained on limited data. Interestingly, with 1M training data, the baseline DP improves significantly and even outperforms SSDS-DP in some splits. Furthermore, one qualitative comparison for the 650K model is presented in Figure \ref{fig:comp_navi}, where SSDS-DP  respects the navigation route whereas the baseline goes out of the navigation, highlighting that scene fusion is more effective in SSDS-DP. 

\begin{table}[ht]
\centering
\caption{Closed-loop nuPlan scores across training-set sizes.}
\label{tab:planning}
\setlength{\tabcolsep}{4.5pt}
\begin{tabular}{llcccccc}
\toprule
& & \multicolumn{3}{c}{Reactive} & \multicolumn{3}{c}{Non-Reactive} \\
\cmidrule(lr){3-5}\cmidrule(lr){6-8}
Data & Method & Val14 $\uparrow$ & T14-hard $\uparrow$ & Test14 $\uparrow$ & Val14 $\uparrow$  & T14-hard $\uparrow$ & Test14 $\uparrow$ \\
\midrule
\multirow{2}{*}{250K} & DP & 73.58 & 55.88 & 76.55 & 84.96 & 71.88 & 87.33 \\
 & SSDS-DP & \textbf{75.85} & \textbf{62.68} & \textbf{77.53} & \textbf{86.25} & \textbf{72.11} & \textbf{88.31} \\
\midrule
\multirow{2}{*}{650K} & DP & 78.69 & 51.89 & 68.74 & \textbf{87.45} & 68.45 & 79.63 \\
 & SSDS-DP & \textbf{78.96} & \textbf{66.29} & \textbf{80.06} & 86.58 & \textbf{69.86} & \textbf{85.72} \\
\midrule
\multirow{2}{*}{1M} & DP & 82.82 & \textbf{69.26} & \textbf{82.89} & \textbf{89.88} & 75.38 & 89.29 \\
 & SSDS-DP & \textbf{83.09} & 65.93 & 79.38 & 87.12 & \textbf{75.85} & \textbf{89.46} \\
\bottomrule
\end{tabular}
\end{table}

\subsection{Guidance: DAPSE}
To highlight the effectiveness of the proposed DAPSE sampling method, we take a simple case of speed maintenance with the following guidance function: 
\begin{equation}
\label{eq:speed_maintain_energy}
\mathcal{E}_{\text{target\_speed}} = \max \left( \frac{\mathrm{d}x_{\text{ego}}^{\tau}}{\mathrm{d}\tau} - v_{\text{low}}, 0 \right)^{2} + \max \left( v_{\text{high}} - \frac{\mathrm{d}x_{\text{ego}}^{\tau}}{\mathrm{d}\tau}, 0 \right)^{2}
\end{equation}

\begin{figure}[!ht]
  \begin{center}
    \includegraphics[width=0.9\textwidth]{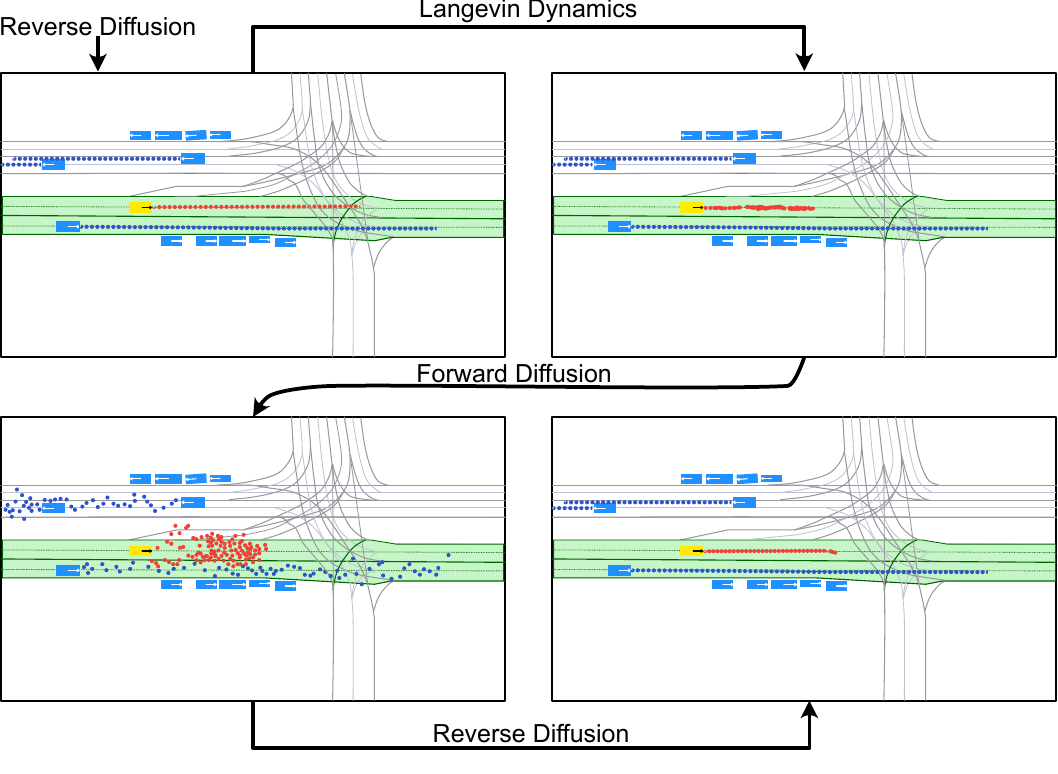}
  \end{center}
  \caption{Speed maintenance with DAPSE.}
  \label{fig:dapse_speed}
\end{figure}

The different steps of this method are presented in Figure \ref{fig:dapse_speed}. To elaborate, after performing the traditional reverse diffusion, the length of the trajectory of ego, in the top left of Figure \ref{fig:dapse_speed} is very long, indicating high speed. After performing the energy update with Langevin dynamics \cite{langevin}, the trajectory is shrunk but becomes more jittery. In the next step, noise is added, and reverse diffusion is performed, and as expected in Figure \ref{fig:dapse_speed} bottom right, the length of the trajectory is smaller compared to the original unconditional sample.

\subsection{Generation: stress-testing planners on adversarial scenarios}
 

The generated scenarios cover four interaction types using different guidance combinations: (1) \textbf{Cut-in}, using lane-change guidance to encourage lateral merging; (2) \textbf{Lead-Agent brake}, combining braking and keep-in-lane guidance to induce longitudinal conflicts while preventing unrealistic lateral deviations; (3) \textbf{Combined cut-in and braking}, sequentially composing cut-in and lead-agent brake to create more challenging interactions; and (4) \textbf{Intersection}, combining TTC guidance with drivable-area compliance to encourage safety-critical interactions while maintaining road compliance.

\begin{table}[!ht]
\centering
\small 

\caption{Closed-loop evaluation on 25 selected Val14 scenarios with combined cut-in and braking scenario. No-Adv: scenarios without any adversary; 
Adv: scenarios with one adversary.}
\label{tab:metric_results}

\setlength{\tabcolsep}{3pt}
\renewcommand{\arraystretch}{1.15}

\begin{tabular}{llcccc|c}
\toprule
\textbf{Scenario} & \textbf{Planner} 
& \begin{tabular}{c}
\textbf{nuPlan} \\
\textbf{Score} \\
{\scriptsize $\uparrow$}
\end{tabular}
& \begin{tabular}{c}
\textbf{TTC} \\
\textbf{Bound} \\
{\scriptsize (\%)$\uparrow$}
\end{tabular}
& \begin{tabular}{c}
\textbf{Ego-Fault} \\
\textbf{Collision} \\
{\scriptsize (\%)$\downarrow$}
\end{tabular}
& \begin{tabular}{c}
\textbf{Ego} \\
\textbf{Comfort} \\
{\scriptsize (\%)$\uparrow$}
\end{tabular}
& \begin{tabular}{c}
\textbf{Realism } \\
\textbf{deviation}\\
{\scriptsize $\downarrow$}
\end{tabular}
\\
\midrule

\multirow{2}{*}{No-Adv} 
    & DP & 80.49 & 91 & 4  & 70 & --- \\
    & SSDS-DP &  \textbf{84.65} &  \textbf{96} & 4 & \textbf{83} & --- \\ 

\midrule

\multirow{2}{*}{Adv} 
    & DP & \textbf{67.29} & \textbf{87} & \textbf{23}  & 61 & 0.38 \\
    & SSDS-DP  & 53.96 & 70 & 26 & \textbf{78} & 0.36 \\

\bottomrule
\end{tabular}
\label{tab:scenario}
\end{table}

Table~\ref{tab:scenario} presents an evaluation of the two diffusion-based planners under the combined cut-in and braking scenario. Under adversarial guidance, both planners experience substantial degradation in closed-loop performance, with SSDS-DP showing a larger drop compared to DP. Both planners exhibit increased ego-at-fault collisions and reduced TTC-in-bound and comfort scores, indicating that the generated interactions effectively expose safety-critical weaknesses. Following~\cite{zhong2023guided}, we additionally report the realism deviation for the adversarial scenario, which measures how closely the adversary's speed, jerk and longitudinal/lateral accelerations match nominal driving behavior. The resulting values are comparable to those reported by SafeSim~\cite{chang2024safesim} under the same metric, suggesting the generated behaviors remain physically plausible despite substantially degrading planner performance. Although SSDS-DP performs better under nominal conditions, it degrades more under adversarial scenarios.

\begin{figure}[!ht]
  \begin{center}
    \includegraphics[width=0.95\textwidth]{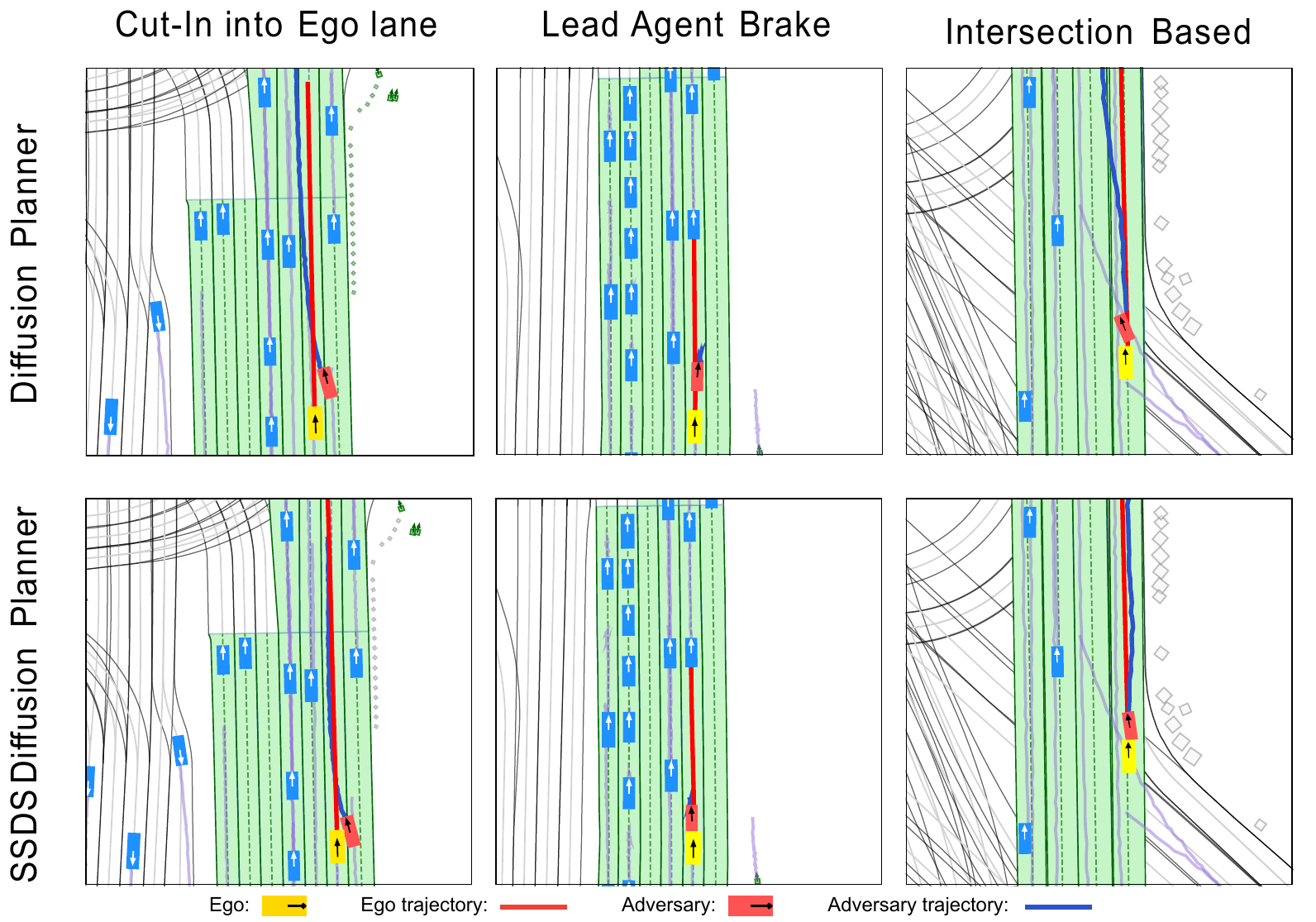}
  \end{center}
  \caption{Comparison between DP and SSDS-DP (1M) on generated adversarial scenarios at the same simulation timesteps.}
  \label{fig:comp_adv_scenario}
\end{figure}

Figure~\ref{fig:comp_adv_scenario} shows a qualitative comparison of the closed-loop responses of DP and SSDS-DP across three different adversarial scenarios. In cut-in scenarios with a limited gap, SSDS-DP often fails to react sufficiently early, resulting in collisions, whereas DP occasionally performs stronger braking responses and avoids the collision. In braking scenarios, DP is able to delay the collision, indicating a more reactive longitudinal response, whereas SSDS-DP does not always apply timely deceleration when the lead vehicle slows, leading to front collisions. Across both scenario types, DP consistently reacts earlier and brakes more decisively than SSDS-DP, mirroring the gap observed in the quantitative results. Notably, in neither scenario type does either planner attempt a lateral evasive maneuver such as overtaking, relying purely on longitudinal deceleration. This limitation becomes decisive in intersection scenarios, where both planners struggle equally, since braking alone cannot resolve an imminent crossing conflict.





The experimental results across planning and scenario generation validate the dual-use capability of our unified diffusion foundation and can be summarised as below:

\begin{itemize}
    \item \textbf{High-fidelity planning via SSDS-DP:} Our SSDS-DP significantly improves context fusion, yielding massive performance leaps in highly interactive, reactive environments (up to $+14.4$ points on Test14-hard) and demonstrating superior data efficiency. However, at 1M training scenarios, there is no clear winner, and we identify multi-seed benchmarking and scaling beyond 1M trajectories as valuable directions for future work.
    
    \item \textbf{Closing the loop via adversarial generation:} By leveraging guided diffusion for scenario generation, our approach enables controllable generation of safety-critical long-tail scenarios, including aggressive cut-ins and braking interactions. In closed-loop evaluation, these scenarios effectively challenge planners that perform strongly under nominal conditions, exposing failure modes while maintaining realistic agent behaviors with low realism deviation.
\end{itemize}
Crucially, the finding that the stronger planner (SSDS-DP) degrades more under generated long-tail scenarios highlights that \emph{standard-benchmark superiority does not guarantee adversarial robustness}. By providing both the planner and the scenario generator within a unified diffusion framework, our work enables systematic identification and stress-testing of complex multi-agent edge cases.

\section{Conclusion}
\label{sec:conclusion}
This paper presented a unified, dual-use foundation for autonomous driving using a single pretrained diffusion model. First, we introduced the SSDS-DP to improve interactive planning and proposed DAPSE, a training-free framework for zero-shot safety steering. Second, we extended the diffusion framework to controllable long-tail scenario generation, enabling realistic closed-loop evaluation scenarios that expose critical planner failure modes. Our analysis reveals that, despite strong nominal benchmark performance, learned planners can struggle with complex multi-agent interactions by relying primarily on reactive braking rather than proactive evasive maneuvers. These results demonstrate that a single learned traffic prior can simultaneously serve as an effective planner and a rigorous evaluator in the autonomous vehicle development loop. Future work will explore other energy functions (e.g., collision avoidance) with the DAPSE framework and quantitatively evaluate its effectiveness (e.g., collision rate, TTC).

 
\section*{Acknowledgments} 
This work was partially supported by the Wallenberg AI, Autonomous Systems and Software Program (WASP) funded by the Knut and Alice Wallenberg Foundation.
\bibliographystyle{splncs04}
\bibliography{main}
\end{document}